\documentclass[runningheads]{llncs}
\usepackage{makeidx}
\usepackage{graphicx}
\usepackage{amsmath,amssymb} 
\usepackage{color}

\newcommand\todo[1]{}

\usepackage{hyperref}
\usepackage[capitalise]{cleveref}
\usepackage{tikz}
\usepackage{pgfplots}
\usepackage{pgfplotstable}
\usepackage{footnote}

\hypersetup{hidelinks=true}

\begin{document}
	\pagestyle{headings}
	\mainmatter

	\title{Improved Semantic Stixels via \\ Multimodal Sensor Fusion}

	\titlerunning{Improved Semantic Stixels}
	\authorrunning{Florian Piewak, Peter Pinggera, et al.}
	\author{Florian Piewak\thanks{Corresponding author (\email{florian.piewak@daimler.com)}}\inst{1,2} \and
	Peter Pinggera\inst{1} \and
	Markus Enzweiler\inst{1} \and \\
	David Pfeiffer\thanks{Contributed while with Daimler AG}\inst{1} \and
	Marius Z\"ollner\inst{2,3}}
	\institute{Daimler AG, R\&D, Stuttgart, Germany \and
		Karlsruhe Institute of Technology (KIT), Karlsruhe, Germany \and
		Forschungszentrum Informatik (FZI), Karlsruhe, Germany}

\definecolor{road}{RGB}{128, 64, 128}
\definecolor{sidewalk}{RGB}{244, 35, 232}
\definecolor{person}{RGB}{220, 20, 60}
\definecolor{rider}{RGB}{255, 0, 0}
\definecolor{smallVehicle}{RGB}{0, 0, 142}
\definecolor{largeVehicle}{RGB}{0, 0, 70}
\definecolor{twoWheeler}{RGB}{119, 11, 32}
\definecolor{construction}{RGB}{70, 70, 70}
\definecolor{pole}{RGB}{153, 153, 153}
\definecolor{trafficSign}{RGB}{220, 220, 0}
\definecolor{vegetation}{RGB}{107, 142, 35}
\definecolor{terrain}{RGB}{152, 251, 152}
\definecolor{sky}{RGB}{70, 130, 180}

\newcommand\insertLabelRow[2]{\textcolor{#1}{$\bullet$} & \textcolor{#1}{#2} \\ }
\newcommand{\insertLabelColorTable}{
									\scriptsize
									\begin{tabular}{ll}
										\insertLabelRow{road}{road}
										\insertLabelRow{sidewalk}{sidewalk}
										\insertLabelRow{person}{person}
										\insertLabelRow{rider}{rider}
										\insertLabelRow{smallVehicle}{small vehicle}
										\insertLabelRow{largeVehicle}{large vehicle}
										\insertLabelRow{twoWheeler}{two wheeler}
										\insertLabelRow{construction}{construction}
										\insertLabelRow{pole}{pole}
										\insertLabelRow{trafficSign}{traffic sign}
										\insertLabelRow{vegetation}{vegetation}
										\insertLabelRow{terrain}{terrain}
									\end{tabular}
									}

\newcommand\stixelTypeGroundText{\textit{support}}
\newcommand\stixelTypeObjectText{\textit{object}}
\newcommand\stixelTypeSkyText{\textit{sky}}

\newcommand\stixelTypeGround{\mathcal{G}}
\newcommand\stixelTypeObject{\mathcal{O}}
\newcommand\stixelTypeSky{\mathcal{S}}

\newcommand\energyLikelihoodVar{\Theta}
\newcommand\energyPriorVar{\Omega}

\newcommand\sfxLidar{lidar}
\newcommand\sfxCam{cam}

\newcommand\stixels{\textbf{\textit{S}}}
\newcommand\stixelFirst{\textbf{\textit{s}}_{1}}
\newcommand\stixel{\textbf{\textit{s}}_{i}}
\newcommand\stixelLast{\textbf{\textit{s}}_{n}}
\newcommand\measuements{\textbf{\textit{M}}}
\newcommand\measuementsOfStixel{\textbf{\textit{M}}^*_i}
\newcommand\measuementFirst{\textbf{\textit{m}}_{1}}
\newcommand\measuement{\textbf{\textit{m}}_{j}}
\newcommand\measuementLast{\textbf{\textit{m}}_{h}}
\newcommand\measuementsDepth{\textbf{\textit{D}}}
\newcommand\measuementDepthFirst{\textbf{\textit{d}}_{1}}
\newcommand\measuementDepth{\textbf{\textit{d}}_{j}}
\newcommand\measuementDistance{r_{j}}
\newcommand\measuementAngleHor{\alpha_{h_{j}}}
\newcommand\measuementAngleVer{\alpha_{v_{j}}}
\newcommand\measuementDepthLast{\textbf{\textit{d}}_{h}}
\newcommand\measuementDepthPre{\textbf{\textit{d}}_{j-1}}
\newcommand\measuementDepthOther{\textbf{\textit{d}}_{k}}
\newcommand\measuementsLabelCam{\textbf{\textit{L}}_{\sfxCam}}
\newcommand\measuementLabelCamFirst{\textbf{\textit{l}}_{\sfxCam_1}}
\newcommand\measuementLabelCam{\textbf{\textit{l}}_{\sfxCam_j}}
\newcommand\measuementLabelCamLast{\textbf{\textit{l}}_{\sfxCam_h}}
\newcommand\measuementsLabelLidar{\textbf{\textit{L}}_{\sfxLidar}}
\newcommand\measuementLabelLidarFirst{\textbf{\textit{l}}_{\sfxLidar_1}}
\newcommand\measuementLabelLidar{\textbf{\textit{l}}_{\sfxLidar_j}}
\newcommand\measuementLabelLidarLast{\textbf{\textit{l}}_{\sfxLidar_h}}

\newcommand\stixelLabel{l}
\newcommand\stixelDistance{r}
\newcommand\stixelType{c}
\newcommand\stixelBottom{b}
\newcommand\stixelTop{t}
\newcommand\stixelPoint{p}

\newcommand{\gradientAngle}{\phi}

\newcommand\probStixelbyMeasurement{P(\stixels | \measuements)}
\newcommand\probMeasurementbyStixel{P(\measuements | \stixels)}
\newcommand\probMeasurement{P(\measuements)}
\newcommand\probStixel{P(\stixels)}
\newcommand\probObject{P_{ob}}

\newcommand\probLikelihoodDist{P_{geo_{dist}}(\measuementDepth | \stixel )}

\newcommand\probLidarOutlier{p_{out}}

\newcommand\energyFunction{E(\dots)}
\newcommand\energyFunctionPosterior{E(\stixels, \measuements)}
\newcommand\energyLikelihood{\energyLikelihoodVar(\stixels, \measuements)}
\newcommand\energyLikelihoodGeometry{\energyLikelihoodVar_{geo}(\stixel, \measuementDepth, \measuementDepthPre)}
\newcommand\energyLikelihoodGeometryDist{\energyLikelihoodVar_{dist}(\stixel, \measuementDepth)}
\newcommand\energyLikelihoodGeometryGrad{\energyLikelihoodVar_{gr}(\stixel, \measuementDepth, \measuementDepthPre)}
\newcommand\energyLikelihoodGeometrySensor{\energyLikelihoodVar_{sens}(\stixel, \measuementDepth)}
\newcommand\energyLikelihoodSemLiDAR{\energyLikelihoodVar_{sem_{\sfxLidar}}(\stixel, \measuementLabelLidar)}
\newcommand\energyLikelihoodSemCam{\energyLikelihoodVar_{sem_{\sfxCam}}(\stixel, \measuementLabelCam)}

\newcommand\energyPrior{\energyPriorVar(\stixels)}
\newcommand\energyPriorModelComplexity{\energyPriorVar_{mc}(\stixels)}
\newcommand\energyPriorModelComplexityFilled{\energyPriorVar_{mc}(\stixelFirst, \dots, \stixelLast)}
\newcommand\energyPriorSegmentationConsistency{\energyPriorVar_{sc}(\stixels)}
\newcommand\energyPriorSegmentationConsistencyFirst{\energyPriorVar_{sc_{1st}}(\stixels)}
\newcommand\energyPriorSegmentationConsistencyFirstFilled{\energyPriorVar_{sc_{1st}}(\stixelFirst, \dots, \stixelLast)}
\newcommand\energyPriorSegmentationConsistencyLast{\energyPriorVar_{sc_{last}}(\stixels)}
\newcommand\energyPriorSegmentationConsistencyLastFilled{\energyPriorVar_{sc_{last}}(\stixelFirst, \dots, \stixelLast)}
\newcommand\energyPriorSegmentationConsistencyOrder{\energyPriorVar_{sc_{ord}}(\stixels)}
\newcommand\energyPriorSegmentationConsistencyOrderFilled{\energyPriorVar_{sc_{ord}}(\stixelFirst, \dots, \stixelLast)}
\newcommand\energyPriorSegmentationConsistencyOrderOneStixel{\energyPriorVar_{i,sc_{ord}}(\stixel)}
\newcommand\energyPriorSegmentationConsistencyConnect{\energyPriorVar_{sc_{con}}(\stixels)}
\newcommand\energyPriorSegmentationConsistencyConnectFilled{\energyPriorVar_{sc_{con}}(\stixelFirst, \dots, \stixelLast)}
\newcommand\energyPriorSegmentationConsistencyConnectTwoStixels[2]{\energyPriorVar_{i,sc_{con}}(\textbf{\textit{s}}_{#1}, \textbf{\textit{s}}_{#2})}
\newcommand\energyPriorGeometricConsistency{\energyPriorVar_{gc}(\stixels)}
\newcommand\energyPriorGeometricConsistencyFilled{\energyPriorVar_{gc}(\stixelFirst, \dots, \stixelLast)}
\newcommand\energyPriorGeometricConsistencyTwoStixels[2]{\energyPriorVar_i,{gc}(\textbf{\textit{s}}_{#1}, \textbf{\textit{s}}_{#2})}

\newcommand\cutCost{\beta_{mc}}
\newcommand\weightLidarGeo{\beta_{geo_{\sfxLidar}}}
\newcommand\weightLidarSem{\beta_{sem_{\sfxLidar}}}
\newcommand\weightCamera{\beta_{sem_{\sfxCam}}}

\newcommand\gradSteepness{\beta_{gr,steep}}
\newcommand\gradShift{\beta_{gr,shift}}
\newcommand\sensorModelScale{\beta_{sens,scale}}
\newcommand\sensorModelShift{\beta_{sens,shift}}

\newcommand\eqDot{\text{ .}}
\newcommand\eqCom{\text{ ,}}

\newcommand\insertplot[4]{
	\pgfplotstableread[col sep=comma]{#1}\datatable
	\foreach \n in {#2} {
		\addplot+ [smooth, #3] table [y index={\n}, x index={#4}] {\datatable};
		\pgfplotstablegetcolumnnamebyindex{\n}\of{\datatable}\to{\colname}
		\addlegendentryexpanded[]{\colname}
	}
}

\newcommand\weightingPlotCam[1]{
	\begin{tikzpicture}[scale=0.9]
	\newcount\xIndex
	\xIndex=3
	\begin{axis}[
	width=\linewidth,
	axis y line*=left,
	xmin=0, xmax=5,
	ymin=45, ymax=75,
	xlabel={Camera Semantic Weight},
	x label style={anchor=near ticklabel, below=0mm},
	xtick={0,1,2,3,4,5},
	scaled ticks=false,
	x tick label style={rotate=90,anchor=east, /pgf/number format/fixed},
	y tick label style={/pgf/number format/fixed, /pgf/number format/fixed zerofill, /pgf/number format/precision=1},
	line width = 1.0pt,
	no markers,
	mark options={scale=0.1},
	restrict y to domain=0:100,
	restrict x to domain=0:10,
	width=\textwidth,
	]
	\insertplot{#1}{12}{blue}{\xIndex} 
	\insertplot{#1}{28}{red}{\xIndex} 
	
	\legend{}
	
	\end{axis}
	
	\begin{axis}[
		width=\linewidth,
		axis y line*=right,
		axis x line=none,
		xmin=0, xmax=5,
		ymin=0.5, ymax=1.5,
		ylabel={\textcolor{brown}{Outlier Rate} in $\%$},
		ylabel near ticks,
		xtick={0,1,2,3,4,5},
		scaled ticks=false,
		y tick label style={/pgf/number format/fixed, /pgf/number format/fixed zerofill, /pgf/number format/precision=1},
		line width = 1.0pt,
		no markers,
		mark options={scale=1.1},
		restrict y to domain=0:100,
		restrict x to domain=0:10,
		width=\textwidth,
		]
		\insertplot{#1}{16}{brown}{\xIndex} 
		
		\legend{}

		\end{axis}
	
	\end{tikzpicture}
}

\newcommand\weightingPlotLiDAR[1]{
	\begin{tikzpicture}[scale=0.9]
	\newcount\xIndex
	\xIndex=2
	\begin{axis}[
	width=\linewidth,
	axis y line*=left,
	xmin=0, xmax=5,
	ymin=45, ymax=75,
	xlabel={LiDAR Semantic Weight},
	ylabel={\textcolor{blue}{Compression Rate} / \textcolor{red}{IoU} in $\%$},
	x label style={anchor=near ticklabel, below=0mm},
	xtick={0,1,2,3,4,5},
	scaled ticks=false,
	x tick label style={rotate=90,anchor=east, /pgf/number format/fixed},
	y tick label style={/pgf/number format/fixed, /pgf/number format/fixed zerofill, /pgf/number format/precision=1},
	line width = 1.0pt,
	no markers,
	mark options={scale=0.1},
	restrict y to domain=0:100,
	restrict x to domain=0:10,
	width=\textwidth,
	]
	\insertplot{#1}{12}{blue}{\xIndex} 
	\insertplot{#1}{28}{red}{\xIndex} 
	
	\legend{}
	
	\end{axis}
	
	\begin{axis}[
		width=\linewidth,
		axis y line*=right,
		axis x line=none,
		xmin=0, xmax=5,
		ymin=0.5, ymax=1.5,
		ylabel near ticks,
		xtick={0,1,2,3,4,5},
		scaled ticks=false,
		y tick label style={/pgf/number format/fixed, /pgf/number format/fixed zerofill, /pgf/number format/precision=1},
		line width = 1.0pt,
		no markers,
		mark options={scale=1.1},
		restrict y to domain=0:100,
		restrict x to domain=0:10,
		width=\textwidth,
		]
		\insertplot{#1}{16}{brown}{\xIndex} 
		
		\legend{}

		\end{axis}
	
	\end{tikzpicture}
}

	\maketitle

	\begin{abstract}
This paper presents a compact and accurate representation of 3D scenes that are 
observed by a LiDAR sensor and a monocular camera.
The proposed method is based on the well-established Stixel model originally developed for stereo vision applications. We extend this Stixel concept
to incorporate data from multiple sensor modalities.
The resulting mid-level fusion scheme takes full advantage of 
the geometric accuracy of LiDAR measurements
as well as the high resolution and semantic detail of RGB images.
The obtained environment model provides a geometrically and
semantically consistent representation of the 3D scene at a
significantly reduced amount of data while minimizing information loss at the same time. 
Since the different sensor modalities are considered as input to a joint optimization problem,
the solution is obtained with only minor computational overhead.
We demonstrate the effectiveness of the proposed multimodal Stixel algorithm on a
manually annotated ground truth dataset.
Our results indicate that the proposed mid-level fusion of LiDAR and camera data improves both the geometric and semantic accuracy
of the Stixel model significantly while reducing the computational overhead as well as the amount of generated data in comparison to using a single modality on its own.
\end{abstract}

	\section{Introduction}
	\label{sec:intorduction}

	Research on autonomous vehicles
has attracted a large amount of attention in recent years,
mainly sparked by the complexity of the problem and the drive to transform the mobility space. One key to success are powerful environment perception systems that allow  autonomous systems to understand and act within a human-designed environment. Stringent requirements regarding accuracy, availability, and safety have led to the use of sensor suites that incorporate complimentary sensor types such as camera, LiDAR, and RADAR. Each sensor modality needs to leverage its specific strengths to  contribute to a holistic picture of the environment.  

\begin{figure}[t]
	\centering
	\begin{minipage}{.8\textwidth}
		\centering
		\includegraphics[width=0.8\linewidth]{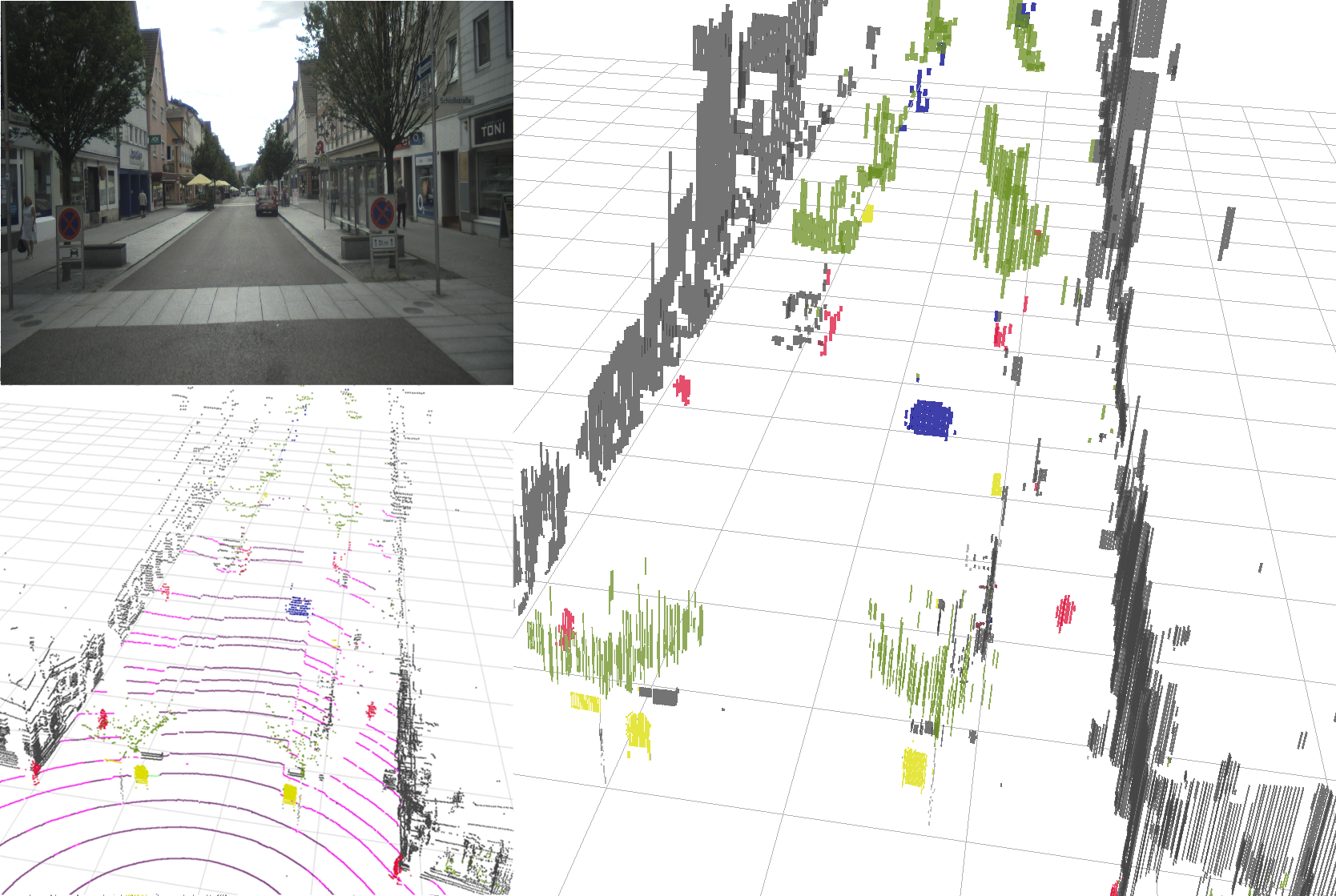}
	\end{minipage}%
	\begin{minipage}{.2\textwidth}
		\insertLabelColorTable
	\end{minipage}
	\caption{Example of a multimodal Stixel scene (right) generated based on a camera image (top left) and a semantic point cloud (bottom left). Note that only \stixelTypeObjectText{} Stixels are visualised. The colors correspond to the Cityscapes semantic class color coding \cite{Cordts2016}.}
	\label{fig:semanticExample}
\end{figure}

The sensor output usually involves quantities that are derived from raw measurements, such as detailed semantics \cite{Cordts2017,Piewak2018}
or object instance knowledge \cite{Yang2016,Zhou2017}.
The different representations provided by the various sensor types
are typically fused into an integrated environment model,
for example an occupancy grid map \cite{Nuss2016},
to successfully tackle high-level tasks such as object
tracking \cite{Vu2011} and path planning \cite{Bai2015}.

Fusing the massive amounts of data provided by multiple different sensors 
represents a significant challenge in a real-time application. As a way out, mid-level data representations have been proposed that
reduce the amount of sensor data but retain the underlying information at the same time. A prime example of such a mid-level representation is the so-called
Stixel-World \cite{Badino2009,Pfeiffer2011,Schneider2016,Cordts2017a,Hernandez-Juarez2017} 
that provides a compact, yet geometrically and semantically consistent
model of the observed environment. Thereby a 3D scene is represented by a set of narrow vertical segments, the Stixels, which are described individually by their vertical extent, geometric surface, and semantic label. 
The Stixel concept was originally applied to stereo camera data,
where the segmentation is primarily based on dense disparity data as well as
pixel-level semantics obtained from a deep neural network~\cite{Schneider2016,Cordts2017a,Hernandez-Juarez2017}. 

In this paper, we propose to transfer the Stixel concept into the LiDAR domain to develop a compact and robust mid-level representation for 3D point clouds. Moreover, we extend the Stixel-World to a multimodal representation by incorporating both camera and LiDAR sensor data into the model. 
The specific combination of the high resolution and semantic detail of RGB imagery with the supreme distance accuracy of LiDAR data in the multimodal Stixel-World results in a very powerful environment representation that outperforms the state-of-the-art (see \cref{fig:semanticExample}). Our main contributions can be summarized as follows:
\begin{itemize}
	\item A compact and robust mid-level representation for semantic LiDAR point clouds based on the Stixel-World
	\item A multimodal fusion approach integrated into the proposed mid-level representation
	\item A detailed performance analysis and quantitative evaluation of the proposed methods
\end{itemize}

\begin{figure}[t]
	\centering
	\includegraphics[width=1.0\linewidth]{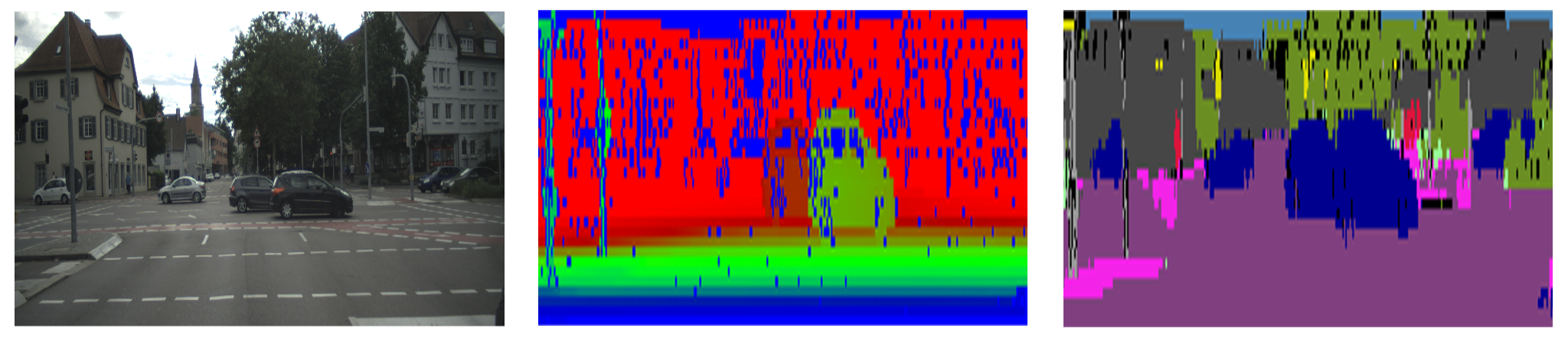}
	\caption{Example of a multimodal Stixel scene (right, colors = Cityscapes semantic class \cite{Cordts2016}, each image column is separated into multiple semantic Stixels) and the corresponding LiDAR distance image (center, blue = close, red = far) projected to a cylindrical view. The corresponding camera image is shown on the left.}
	\label{fig:cylindricalImage}
\end{figure}

	\section{Related Work}
	\label{sec:relatedWork}

	The multimodal Stixel approach presented in this paper combines LiDAR distance
measurements with the point-wise semantic labeling information obtained from both LiDAR and a monocular camera. We relate our approach to three
different categories of existing work: semantic labeling, sensor fusion, and compact mid-level data representations.

First, semantic labeling describes a range of techniques for
the measurement-wise (e.g. pixel-wise) assignment of object class or object type.
The topic has been well explored within the camera domain \cite{Garcia-Garcia2017,Long2015,Cordts2017,Sankaranarayanan2017}.
In contrast, semantic labeling for 3D point clouds is a
relatively recent topic \cite{Qi2017,Riegler2016}, which has mainly been studied on
indoor \cite{Armeni2017,Dai2017} or
stationary outdoor datasets \cite{Hackel2017}.
Within road scenarios,
Wu et al. \cite{Wu2017} introduced a Fully
Convolutional Neural Network (FCN) approach
based on the SqueezeNet architecture \cite{Iandola2016} for semantic labeling
of vehicles, pedestrian and cyclists within 3D LiDAR point cloud data.
A 2D cylindrical projection of the point cloud (see \cref{fig:cylindricalImage})
is applied, enabling the application of efficient image-based filter kernels.
Piewak et al. \cite{Piewak2018} extend this concept and
propose an improved network architecture which is able to perform high-quality semantic labeling
of a 3D point cloud based on 13 classes similar to the Cityscapes Benchmark
suite \cite{Cordts2016}. 
As the multimodal Stixel approach proposed in this paper utilizes semantics from
both LiDAR and camera data,
we apply the approach of \cite{Piewak2018}
to directly extract the detailed point-wise semantics from LiDAR data. This results in a class representation similar to the camera domain,
where we make use of the efficient FCN architecture
described by Cordts et al. \cite{Cordts2017}.

Second, different fusion strategies can be applied to  the multimodal data of various sensors. Several approaches perform so-called low-level fusion by directly combining
the raw data to obtain a joint sensor representation,
which is then used for object detection \cite{Gupta2014} or
semantic labeling \cite{Muller2014}. 
A different method commonly used within
the autonomous driving context is high-level fusion
\cite{Nuss2016}, where the sensor data is processed independently and
the results are later combined on a more abstract level.
In this paper, we present a novel fusion concept which integrates
the sensor data on mid-level,
reducing the data volume while minimizing information loss.
This representation can further be integrated into a more abstract
environment model such as an occupancy grid \cite{Nuss2016}.

Third, the presented multimodal Stixel approach
is closely related to other compact mid-level representations in terms of the output data format. 
In particular, we refer to the Stixel-World obtained from camera imagery,
which has successfully been applied 
with \cite{Benenson2012,Pfeiffer2011,Liu2015}
and without \cite{Benenson2012,Levi2015} the use of stereoscopic depth information.
The integration of camera-based semantic labeling information
into the Stixel generation process was presented in \cite{Schneider2016,Cordts2017,Hernandez-Juarez2017},
thereby further improving robustness and promoting the semantic consistency of the result.
The Stixel concept has also been adapted to other image-based sensor techniques,
for example to a camera-based infrared depth sensor as shown in \cite{Martinez2017}.
Forsberg \cite{Forsberg2018} makes use of a LiDAR scanner to obtain depth
information for the Stixel generation process. Similarly to an early idea in \cite{Pfeiffer2011},
the LiDAR point cloud is simply projected into the camera image to replace the original
dense disparity information with the sparse LiDAR-based depth measurements.
In contrast, we employ a LiDAR-specific sensor model that is particularly tailored to exploit the superior geometric accuracy of the LiDAR sensor over a stereo camera. 
Finally, we integrate semantics from both
LiDAR and camera data into the Stixel generation
process to obtain a high-quality, comprehensive mid-level 3D representation of the environment.

	\section{Method}
	\label{sec:method}

	The proposed Stixel model is inspired by the
stereoscopic camera approaches of
\cite{Pfeiffer2011,Schneider2016,Cordts2017a}.
After a general definition of the Stixel representation,
we describe the transfer of the Stixel model to the LiDAR domain as well as the adapted Stixel generation process.

\subsection{Stixel Definition}

\begin{figure}
	\centering
	\includegraphics[width=0.7\linewidth]{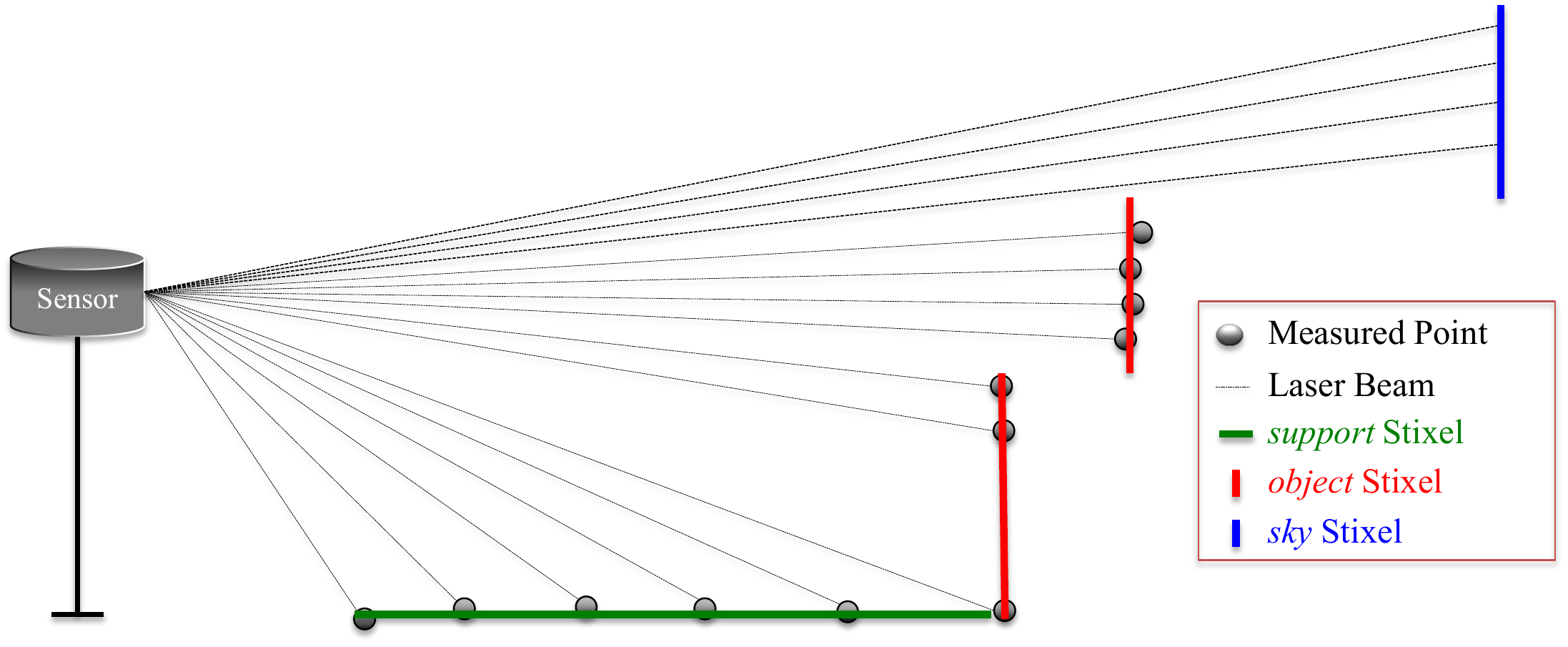}
	\caption{Exemplary Stixel extraction based on a vertical LiDAR scan column.}
	\label{fig:stixelExample}
\end{figure}

Stixels are segments which represent sensor data
in a compact fashion while retaining the underlying semantic and geometric
properties. Generally, the segmentation of an image represents a 2D optimization problem which is
challenging to solve in a real-time environment. Instead, Stixels are optimized column-wise,
which reduces the optimization task to a 1D problem that can be efficiently solved via dynamic programming \cite{Pfeiffer2011}. As a result,
each column is separated into
rectangular stick-like segments $\stixels$ called
Stixels. Within the LiDAR domain, we represent the input data as an 
ordered set of columns of the LiDAR scan, obtained
from a cylindrical projection of the 3D measurements onto a 2D grid, as shown in~\cref{fig:cylindricalImage}.
Each Stixel $\stixel = ( \stixelBottom,
\stixelTop, \stixelDistance, \stixelLabel,
\stixelType )$ is represented by the bottom row
index $\stixelBottom$ and the top row index $\stixelTop$,
describing its vertical extent
with regard to the vertically
ordered measurements $\measuements =
(\measuementFirst, \dots, \measuementLast)$. Additionally, each Stixel has a semantic label
$\stixelLabel$, a structural class
$\stixelType$, and a distance $\stixelDistance$ to the
sensor or the ideal ground plane (depending on
the structural class $\stixelType$). There are three different Stixel
structural classes, i.e. \stixelTypeGroundText{}
($\stixelTypeGround$) for flat regions such as road surface or sidewalk, \stixelTypeObjectText{}
($\stixelTypeObject$) for obstacles such as people or vehicles, and \stixelTypeSkyText{}
($\stixelTypeSky$) for areas without LiDAR measurements, as indicated in~\cref{fig:stixelExample}.

\subsection{Stixel Model}
The vertically ordered (bottom to top) set of measurements
$\measuements =
(\measuementFirst, \dots, \measuementLast)$ is processed column-wise
(see \cref{fig:cylindricalImage}) and 
contains LiDAR depth measurements
$\measuementsDepth = (\measuementDepthFirst,
\dots, \measuementDepthLast)$ 
as well as semantics
from the camera $\measuementsLabelCam =
(\measuementLabelCamFirst, \dots, \measuementLabelCamLast)$
and the LiDAR $\measuementsLabelLidar =
(\measuementLabelLidarFirst, \dots, \measuementLabelLidarLast)$, respectively. The
extraction of semantics from the LiDAR
is done using the LiLaNet architecture of 
\cite{Piewak2018}. The semantic
information of the camera is associated to the 3D
LiDAR points based on the so-called Autolabeling technique
\cite{Piewak2018}, which
projects the LiDAR points into the 
image plane in order to associate the semantics
provided by a state-of-the-art image-based FCN to each point.

Based on this definition the posterior distribution
$\probStixelbyMeasurement$ of the Stixels $\stixels$ given the
measurements $\measuements$ of a column is defined
using the likelihood $\probMeasurementbyStixel$ as well
as the prior $\probStixel$ as
\begin{equation}
	\probStixelbyMeasurement = \frac{\probMeasurementbyStixel \cdot \probStixel}{\probMeasurement} \eqDot
	\label{eq:posteriorStixel}
\end{equation}

Here, the Stixels $\stixels = (\stixelFirst,
\dots, \stixelLast)$ are vertically ordered in accordance with
the measurement vector $\measuements$. Formulating the
posterior distribution in the log-domain yields
\begin{equation}
	\probStixelbyMeasurement = e ^{- \energyFunctionPosterior} \eqCom
	\label{eq:posteriorStixelLogDomain}
\end{equation}
where $\energyFunctionPosterior$ represents an energy function
similar to \cite{Cordts2017a}, defined as
\begin{equation}
	\energyFunctionPosterior = \energyLikelihood + \energyPrior - \log ( \probMeasurement ) \eqDot
	\label{eq:posteriorEnergy}
\end{equation}
Note that $\energyLikelihood$ represents the data likelihood,
$\energyPrior$ the segmentation prior, and $\probMeasurement$ a
normalizing constant.
In contrast to camera-based Stixel applications, as discussed in \cref{sec:relatedWork},
the proposed approach puts forward a LiDAR-specific sensor model to better integrate the accurate LiDAR geometry
into the Stixel-World. This will be discussed within the next subsections.

\subsubsection{Prior}
The prior $\energyPrior$ puts constraints on the Stixel
model in terms of model complexity and segmentation
consistency with
\begin{equation}
	\energyPrior = \energyPriorModelComplexity + \energyPriorSegmentationConsistency \eqDot
	\label{eq:priorEnergy}
\end{equation}
The model complexity term $\energyPriorModelComplexity$ describes the trade-off between the
compactness and the accuracy of the representation. The segmentation consistency $\energyPriorSegmentationConsistency$ governs hard
constraints on the Stixels concerning the relation of
Stixels within a column. The formulation of these prior terms does not depend on the LiDAR measurements, similar to existing Stixel approaches in the camera domain. For further details, the reader is referred to \cite{Cordts2017a}.

\subsubsection{Data Likelihood}

The data likelihood represents the matching quality of the
measurements $\measuements$ to a given set of Stixels $\stixels$,
considering three different data modalities:
LiDAR geometry, LiDAR semantics, and camera semantics:
\begin{equation}
\begin{split}
	\energyLikelihood = \sum\limits_{\stixel \in \stixels} \sum\limits_{\measuement \in \measuementsOfStixel} &\weightLidarGeo \energyLikelihoodGeometry \\
	&+ \weightLidarSem \energyLikelihoodSemLiDAR \\
	&+ \weightCamera \energyLikelihoodSemCam \eqDot
	\label{eq:likelihoodEnergy}
\end{split}
\end{equation}
Here $\measuementsOfStixel$ represents a subset of the
measurements $\measuements$ associated to a specific Stixel $\stixel$.
The parameters $\weightLidarGeo$, $\weightLidarSem$, and
$\weightCamera$ represent weighting parameters of each modality,
which are described within this subsection.

\paragraph{LiDAR Geometry}

The LiDAR geometry data likelihood consists of three
elements defined as follows:
\begin{equation}
\begin{split}
	\energyLikelihoodGeometry &= \energyLikelihoodGeometryDist + \energyLikelihoodGeometryGrad \\
	&+ \energyLikelihoodGeometrySensor \eqDot
	\label{eq:likelihoodEnergyGeometry}
\end{split}
\end{equation}
First of all, the relation of a LiDAR depth measurement  $\measuementDepth$ and the Stixel $\stixel$
is given by the term $\energyLikelihoodGeometryDist$.
\todo{if there is space left, insert an image for visualization of the depth likelihood (chrysa) --> maybe for final version}
We represent this data likelihood as a mixture
of a normal distribution, encoding the sensor noise
based on the variance $\sigma$, and an uniform distribution
representing outlier measurements with an outlier rate of  $\probLidarOutlier$ similar to~\cite{Cordts2017a}.

%
%

In addition to the common depth likelihood definition $\energyLikelihoodGeometryDist$,
two additional likelihood terms are defined
to take advantage of LiDAR-specific measurement properties:
a ground term $\energyLikelihoodGeometryGrad$ and a
sensor term $\energyLikelihoodGeometrySensor$.
The ground term assesses
the consistency of the data with an assumed ground model, based on 
the gradient between two measurements
\begin{equation}
	\gradientAngle(\measuementDepth,\measuementDepthOther) = \arctan\left(\frac{\Delta z_{jk}}{d_{ground_{jk}}}\right)
	= \arctan\left(\frac{z_k - z_j}{\sqrt{x_k^{2} + y_k^{2}} - \sqrt{x_j^{2} + y_j^{2}}}\right) \eqDot
	\label{eq:lidarGrandient}
\end{equation}
Note that a geometric LiDAR measurement
$\measuementDepth = (\measuementDistance, \measuementAngleHor,
\measuementAngleVer)$ is represented using polar coordinates
and consists of a measured distance $\measuementDistance$,
a horizontal angle $\measuementAngleHor$, and a vertical
angle $\measuementAngleVer$. Based on these polar coordinates,
the Cartesian coordinates $(x_j, y_j, z_j)$
are extracted.

The gradient $\gradientAngle$ obtained from the high-quality LiDAR measurements provides structural information of the environment to distinguish between flat surfaces such as ground (low gradient) and obstacles (high gradient). This information is encoded into an object existence probability
using a parametrized hyperbolic tangent as
\begin{equation}
	\probObject(\measuementDepth,\measuementDepthPre) = \frac{1 + \tanh( \gradSteepness (\gradientAngle (\measuementDepth,\measuementDepthPre) - \gradShift) )}{2} \eqDot
	\label{eq:probGrandObject}
\end{equation}
Note that the parameters $\gradSteepness$ and $\gradShift$ adapt the sensitivity of the gradient model. Subsequently, the data likelihood based on the ground model is defined as
\begin{equation}
	\energyLikelihoodGeometryGrad = 
	\begin{cases}
	-\log (1 - \probObject(\measuementDepth,\measuementDepthPre)) &  \text{if } \gradientAngle \text{ is def. and } c_i =\stixelTypeGround\\
	-\log (\probObject(\measuementDepth,\measuementDepthPre)) &  \text{if } \gradientAngle \text{ is def. and } c_i = \stixelTypeObject\\
	0   			&  \text{if } \gradientAngle \text{ is undef. or } c_i = \stixelTypeSky
	\end{cases}
	\eqDot
	\label{eq:costLidarGradGround}
\end{equation}
Note that the data likelihood based on the ground model
is set to zero when the gradient is undefined, which can be caused by missing
reflections of the LiDAR laser light (e.g. if the laser
beam is pointing to the sky).
However, both the vertical and horizontal
angles of the polar coordinate of the so-called invalid
measurement are still available.

In case of an invalid measurement, the data likelihood
based on both the ground model and the depth
matching cannot be processed. For this reason we introduce
the sensor term $\energyLikelihoodGeometrySensor$ to the likelihood formulation,
which is based on the vertical distribution of measurement angles
of the LiDAR sensor. We assume that a \stixelTypeSkyText{}
Stixel is more likely to occur at larger vertical angles, which is encoded
into a parametrized hyperbolic tangent similar to
\cref{eq:probGrandObject} as
\begin{equation}
	P_{\stixelTypeSky}(\alpha_{v_j}) = \frac{1 + \tanh( \sensorModelScale (\alpha_{v_j} - \sensorModelShift)) }{2} \eqDot
\end{equation}
A similar definition is used with regard to small vertical angles and \stixelTypeGroundText{} Stixels by inverting the vertical angle $P_{\stixelTypeGround}(\alpha_{v_j}) = P_{\stixelTypeSky}(-\alpha_{v_j})$.
Consequently, the sensor term contribution for invalid points is defined by
\begin{equation}
	\energyLikelihoodGeometrySensor =
	\begin{cases}
	-\log (P_{\stixelTypeSky}(\alpha_{v_j})) &  \text{if } \measuementDepth \text{ is invalid and } c_i = \stixelTypeSky \\
	-\log (P_{\stixelTypeGround}(\alpha_{v_j})) &  \text{if } \measuementDepth \text{ is invalid and } c_i = \stixelTypeGround \\
	-\log (1 - P_{B}(\alpha_{v_j})) &  \text{if } \measuementDepth \text{ is invalid and } c_i = \stixelTypeObject \\
	\infty &  \text{if } \measuementDepth \text{ is valid and } c_i = \stixelTypeSky \\
	0 &  \text{if } \measuementDepth \text{ is valid and } c_i \in \{\stixelTypeGround, \stixelTypeObject \} \\
	\end{cases}
	\eqCom
	\label{eq:costLidarGradSensor}
\end{equation}
with $P_{B}(\alpha_{v_j}) = P_{\stixelTypeSky}(\alpha_{v_j}) + P_{\stixelTypeGround}(\alpha_{v_j})$. Note that a hard constraint is inserted to prohibit \stixelTypeSkyText{} Stixels resulting from valid measurements.

\paragraph{Semantic Information}
The semantic information obtained from the LiDAR data is utilized in a
similar way as in the Stixel-World of the camera domain.
Each semantic measurement
$\measuementLabelLidar$ holds a probability estimate $P_{\measuementLabelLidar}(l)$
of each class $l$ conditioned on the input data,
which can be obtained from the underlying semantic labeling method.
We make use of the LiLaNet architecture of
\cite{Piewak2018} to compute the point-wise LiDAR-based
semantic information. The definition of the semantic data likelihood
is adapted from \cite{Schneider2016} and
\cite{Cordts2017a} as
\begin{equation}
	\energyLikelihoodSemLiDAR = - \log (P_{\measuementLabelLidar}(l_i)) \eqDot
	\label{eq:lidarSemantic}
\end{equation}
To obtain high resolution semantic information from the camera image, we make use of the efficient FCN architecture described by Cordts et al.
\cite{Cordts2017}. Fusing this information into the proposed multimodal
Stixel approach enables the combination of high resolution camera semantics 
with geometrically accurate information of the
LiDAR. For this purpose we apply the projection technique of
\cite{Piewak2018} to extract the semantic
information of the camera by projecting the LiDAR measurements into
the semantically labeled image. Each LiDAR measurement then
holds additional semantic information from the camera domain $\measuementLabelCam$
which is processed similar to \cref{eq:lidarSemantic} based on the
probability $P_{\measuementLabelCam}(l)$ for each semantic class $l$ with
\begin{equation}
	\energyLikelihoodSemCam = - \log (P_{\measuementLabelCam}(l_i)) \eqDot
	\label{eq:CameraSemantic}
\end{equation}
Note that this definition is independent of the LiDAR-based semantics which enables the extraction of different domain specific semantic classes $l$ from camera and LiDAR.
Especially the camera-based FCN \cite{Cordts2017} extracts more semantic classes based on the higher resolution as well as the larger receptive field as the LiDAR-based FCN \cite{Piewak2018}.
Hence, the domain specific strengths of each sensor modality and the differing object appearance within the LiDAR and the camera are combined to increase the semantic consistency
of the multimodal Stixel result.

\subsection{Stixel Generation}

Based on the proposed Stixel model, Stixels are generated
by finding the maximum-a-posteriori solution of
\cref{eq:posteriorStixel}. This is equal to the
minimization of the energy function given in  \cref{eq:posteriorEnergy}.
Note, that the probability of the measurement $\probMeasurement$
represents a scaling factor which is ignored within the
optimization process. To solve this 1D column-wise optimization process,
a dynamic programming approach is used similar to the original Stixel formulation~(c.f.~\cite{Pfeiffer2011} and \cite{Cordts2017a}).



	\section{Experiments}
	\label{sec:experiments}

	To evaluate our proposed multimodal Stixel model,
we use the manually annotated  dataset of Piewak
et al. \cite{Piewak2018}.
The dataset consists of manually annotated semantic LiDAR point clouds
recorded from a vehicle in various traffic scenarios,
and further includes corresponding image data captured by a front-facing monocular camera.
This enables
both a semantic evaluation of our proposed
method based on the manually annotated semantic LiDAR
data and a geometric evaluation based on the
LiDAR depth data. Due to the sensor configuration within the dataset,
the evaluation is restricted to the area inside the field of view of the camera.
We evaluate various performance metrics on a point-wise
basis to measure the geometric and semantic consistency
as well as the compactness of the model:
\begin{enumerate}
	\item Outlier Rate
	
	A relative distance deviation of the original LiDAR depth measurement
	from the associated Stixel of more than $5\%$ is declared as an outlier.
    Based on this
	formulation the outlier rate is defined as the ratio of
	the number of outliers to the number of total LiDAR points.
	
	\item Intersection over Union (IoU)
	
	Based on the manually annotated semantic ground truth,
	an IoU of the Stixels to the ground truth LiDAR points
	can be calculated similar to \cite{Cordts2016}.
	
	\item Compression Rate
	
	The data compression rate $\theta$ defines the ratio
	between the number of stixels $n_{stixels}$ and number
	of original LiDAR points $n_{points}$ via
	\begin{equation}
		\theta = 1- \frac{n_{stixels}}{n_{points}} \eqDot
	\end{equation}

\end{enumerate}
\begin{figure}[t]
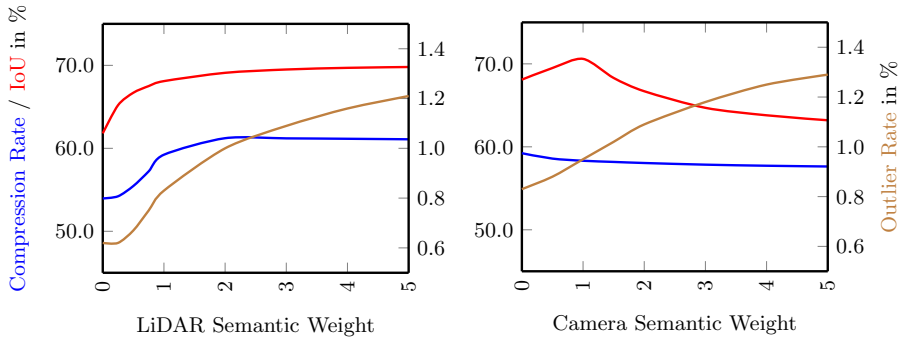

	\centering%
	\begin{minipage}{.5\linewidth}
		\centering
		\weightingPlotLiDAR{data/stixelEvalLiDAR.csv}%
	\end{minipage}%
	\begin{minipage}{.5\linewidth}
		\centering
		\vspace{3mm}
		\weightingPlotCam{data/stixelEvalCam.csv}%
	\end{minipage}
	\caption{Impact of considering the semantic information within the multimodal Stixel model with a constant LiDAR geometry weight ($\weightLidarGeo = 1$). left: Adaption of the LiDAR semantic weight $\weightLidarSem$ based on a deactivated camera semantics ($\weightCamera = 0$). right: Adaption of the camera semantic weight $\weightCamera$ based on a LiDAR semantic weight $\weightLidarSem = 1$}
	\label{fig:lidarWeighting}
\end{figure}
The quantitative results are illustrated in
\cref{fig:lidarWeighting}. First, 
the impact of the LiDAR semantic weight
$\weightLidarSem$  is evaluated while the LiDAR geometry weight
is set to $\weightLidarGeo = 1$ and the camera semantics
is deactivated ($\weightCamera = 0)$. We observe
that the semantic consistency is constantly increasing
with an increase of the LiDAR semantic weight. At
the same time, the compression rate increases as well
as the outlier rate. Putting too much focus on
the semantic input thus reduces the number of individual Stixels
and yields a model purely tuned to the LiDAR semantics.
In turn, consistency with the underlying geometry decreases.

Considering the multimodality in our model by activating
the camera semantics, the compression rate
as well as the outlier rate slightly decreases.
The semantic consistency further
improves until the weighting of the camera semantics
$\weightCamera$ reaches the weighting of
the LiDAR semantics $\weightLidarSem$.
However, the camera semantics on its own reaches a lower IoU
after the transfer to the LiDAR domain (see \cref{tab:overviewReults}). This
demonstrates the potential of our novel multimodal Stixel
approach, which creates a compact, geometrically and semantically
consistent mid-level representation by combining the
advantages of different sensor domains to reach a higher
accuracy than each modality on its own. Our proposed
method of equally weighting the
different modalities represents the best combination
with regard to the semantic consistency as well as a good compromise concerning
the outlier rate and the compression rate. This setup outperforms
the original Stixel-World based on a stereoscopic camera regarding
the geometric and the semantic consistency
of the data representation (see \cref{tab:overviewReults}).

\begin{savenotes}
\begin{table}[t]
	\centering
	\caption{Comparison of the original Stixel World (based on a stereoscopic camera), the different independent Stixel optimization
		modalities, and the combined multimodal
		representation. \todo{pay attention of the footnote}}
	\label{tab:overviewReults}
	\newcolumntype{S}{>{\centering\arraybackslash} m{1.45cm} }
	\begin{tabular}{|c|S|S|S|S|S|}
		\hline
		    & Stereo Camera \cite{Cordts2017a}\footnote{Results of the original Stixel-World (stereo camera) are added for comparison based on \cite{Cordts2017a}. No evaluation is done on our dataset.} & LiDAR Depth only & LiDAR Semantics only & Camera Semantics only & Multi-Modality \\
		\hline
		Outlier Rate in \%     &  6.7 & \textbf{0.62} & 28.8 & 35.3 & 0.95 \\
		\hline
		IoU in \%              & 66.5 & 61.8 & 70.0 & 60.8 & \textbf{70.6} \\
		\hline
		Compression Rate in \% &  \_ & 54.0 & 81.2 & \textbf{85.3} & 58.3 \\
		\hline	
	\end{tabular}
\end{table}
\end{savenotes}

	\section{Conclusion}
	\label{sec:conclusion}

	In this paper, we presented the multimodal Stixel-World,
a Stixel-based environment representation to directly leverage both camera and LiDAR sensor
data. Our design goal
is to jointly represent accurate geometric and semantic
information based on a multi-sensor system within a compact and efficient environment model.
To this end we introduce a LiDAR-specific sensor model that exploits
the geometric accuracy of LiDAR sensors as well as a mid-level fusion
technique to combine valuable semantic information from both camera and LiDAR.
In our experiments we demonstrated the benefits of our multimodal
Stixel-World over unimodal representations in terms of
representation and compression quality by outperforming
the original Stixel-World based on a stereoscopic camera.
Moreover, our presented
multimodal Stixel approach can easily
be extended to other sensor modalities as long
as they can be projected into a common
structured data format.



	\bibliographystyle{splncs03}

\end{document}